\documentclass[10pt,twocolumn,letterpaper]{article}
\usepackage[utf8]{inputenc} 
\usepackage[T1]{fontenc}   
\usepackage{iccv}              

\definecolor{iccvblue}{rgb}{0.21,0.49,0.74}
\usepackage[pagebackref,breaklinks,colorlinks,allcolors=iccvblue]{hyperref}

\usepackage{caption}
\usepackage{subcaption}
\usepackage{graphicx} 
\usepackage[american]{babel}
\usepackage{microtype}
\usepackage{multirow}
\usepackage{booktabs}
\usepackage{algorithm}
\usepackage{algorithmic}

\usepackage{bm}

\usepackage{wrapfig}
\usepackage{pifont}
\usepackage{multirow}
\usepackage[export]{adjustbox}
\usepackage{color}
\usepackage{colortbl}
\usepackage{lipsum}
\usepackage{pifont}       
\usepackage{bbding}    
\usepackage[dvipsnames]{xcolor}
\usepackage{array}

\usepackage[capitalize]{cleveref}
\usepackage{float}
\crefname{section}{Sec.}{Secs.}
\Crefname{section}{Section}{Sections}
\Crefname{table}{Table}{Tables}
\crefname{table}{Tab.}{Tabs.}
\definecolor{red}{RGB}{255,0,0}
\definecolor{blue}{RGB}{0,0,255}
\definecolor{green}{RGB}{0,255,0}
\definecolor{mygray}{gray}{.9}
\definecolor{mygray2}{gray}{.5}
\definecolor{mywarning}{RGB}{233,144,61}
\definecolor{mygreen}{RGB}{93,174,86}
\definecolor{codefunc}{RGB}{73,122,234}

\usepackage[american]{babel}
\usepackage{microtype}

\definecolor{mygreen}{RGB}{0,154,85}
\definecolor{myy}{RGB}{126,95,0}
\definecolor{myred}{RGB}{212,121,116}
\definecolor{myblue}{RGB}{184, 134, 73}
\definecolor{mynewgreen}{RGB}{113,188,169}
\definecolor{mypurple}{RGB}{123,104,238}
\colorlet{R1}{myblue}
\colorlet{R2}{mypurple}
\colorlet{R3}{myred}
\colorlet{R6}{mypurple}
\definecolor{mycite}{RGB}{73,123,184}
\colorlet{cite}{mycite}

\newcommand{\T}{{\top}}
\newcommand{\x}{{\bm x}}

\newcommand{\z}{{\bm z}}

\newcommand{\EPS}{{\bm\epsilon}}
\newcommand{\q}{{\bm{q}}}
\newcommand{\A}{{\bm{a}}}

\def\paperID{5263} 
\def\confName{ICCV}
\def\confYear{2025}

\title{Adversarial-Guided Diffusion for Multimodal LLM Attacks}

\author{
	Chengwei Xia$^1$, Fan Ma$^2$, Ruijie Quan$^3$, Kun Zhan$^{1,\star}$, and Yi Yang$^{2}$\\
	1. School of Information Science and Engineering, Lanzhou University\\
	2. College of Computer Science and Technology, Zhejiang University\\
	3. College of Computing and Data Science, Nanyang Technological University\\
	{\small \url{https://github.com/kunzhan/AGD}}
}

\begin{document}
\maketitle
\begin{abstract}
This paper addresses the challenge of generating adversarial image using a diffusion model to deceive multimodal large language models (MLLMs) into generating the targeted responses, while avoiding significant distortion of the clean image. To address the above challenges, we propose an adversarial-guided diffusion (AGD) approach for adversarial attack MLLMs. We introduce adversarial-guided noise to ensure attack efficacy. A key observation in our design is that, unlike most traditional adversarial attacks which embed high-frequency perturbations directly into the clean image, AGD injects target semantics into the noise component of the reverse diffusion. Since the added noise in a diffusion model spans the entire frequency spectrum, the adversarial signal embedded within it also inherits this full-spectrum property. Importantly, during reverse diffusion, the adversarial image is formed as a linear combination of the clean image and the noise. Thus, when applying defenses such as a simple low-pass filtering, which act independently on each component, the adversarial image within the noise component is less likely to be suppressed, as it is not confined to the high-frequency band. This makes AGD inherently robust to variety defenses. Extensive experiments demonstrate that our AGD outperforms state-of-the-art methods in attack performance as well as in model robustness to some defenses.
\end{abstract}
\section{Introduction}\label{sec:intro}
Multimodal large language models (MLLMs) have achieved superior performance across various tasks~\cite{li2023blip,bao2023one,zhu2023minigpt,liu2024visual,xie2025chain}. However, recent studies have shown that the integration of image modality in MLLM increases their susceptibility to adversarial attacks. In particular, MLLM can be easily misled by adversarial image, which are injected by introducing imperceptible perturbations to the clean image~\cite{zhao2023evaluate,cui2024robustness,luo2024an}. These findings highlight the importance of studying the MLLM adversarial attack.

\begin{figure}[!t]
	\centering
	\includegraphics[width=0.5\textwidth]{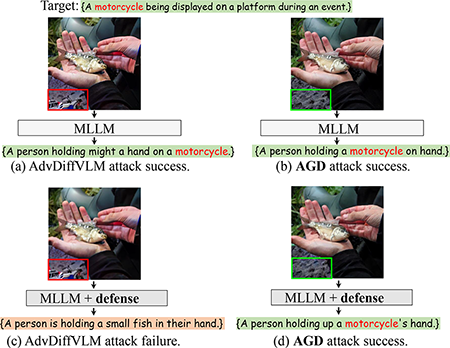}
	\caption{
		AGD injects the target information in a manner that is visually imperceptible, while also maintaining effectiveness against defense mechanisms. (a) Although AdvDiffVLM successfully attacks MLLM, the target (motorcycle) is easily detectable, as shown by the red bounding box.
		(b) In contrast to AdvDiffVLM, AGD successfully attacks MLLM while maintaining imperceptibility. (c) AdvDiffVLM is defended by a defense mechanism~\cite{nie2022DiffPure} and its target still easily detectable. (d) Under the same defense mechanism~\cite{nie2022DiffPure}, AGD not only successfully attacks MLLM but also effectively hides the target.
	}
\label{show-intro}
\end{figure}

It is natural to consider injecting imperceptible perturbations into the high-frequency components of the clean image~\cite{zhao2023evaluate,xie2025chain}. Injecting target information into high-frequency components preserves essential visual content, avoiding significant distortion of the original image~\cite{pei2025diffusion}. However, such perturbations are often vulnerable to simple low-pass filtering defenses, which can effectively remove high-frequency noise.
This leads to suboptimal attack robustness and degraded performance, limiting the utility of these attack methods for evaluating robustness. Most existing diffusion-based adversarial attack~\cite{chen2023advdiffuser,dai2023advdiff,chen2023contentbased,guo2024efficient} methods generate adversarial images with image fusion in all reverse-diffusion process, inducing significant perturbations that make the adversarial image overly noticeable to users, triggering defense mechanisms~\cite{nie2022DiffPure} as shown in Figure~\ref{show-intro}.
In this paper, the adversarial image is generated through a few adversarial-guided diffusion (AGD) steps of the reverse-diffusion process in a text-to-image generative diffusion model. Specifically, AGD guides the injection of target information into the clean image during the final few steps. By fully embedding the target information into the adversarial image, a momentum-based gradient update simulates the iterative denoising process to find the optimal adversarial direction.
AGD ensures attack efficacy while minimizing perturbations. Moreover, AGD embeds targets across the all frequency bands, unlike conventional methods that focus on high-frequency components~\cite{pei2025diffusion}, ensuring that the adversarial image of AGD is less likely to be smoothed out by defense mechanisms, thereby maintaining robustness.


The contributions of this paepr are summarized as follows:
\begin{itemize}
\item We propose an adversarial-guided diffusion (AGD) method to generate adversarial images by injecting targeted semantic information. AGD is effective not only in generating high-quality adversarial image close to the clean image but also in defending against defense mechanism.
\item  Based on the key observation of previous attack, AGD ensures attack robustness while maintaining high fidelity by injecting target into noise, providing a novel perspective for performing targeted adversarial attacks on MLLMs.
\item Experiments conduct on several popular open-source MLLMs demonstrate that AGD outperforms previous state-of-the-art (SOTA) attack methods.
\end{itemize}
\begin{figure*}[!t]
\centering
\includegraphics[width=0.90\linewidth]{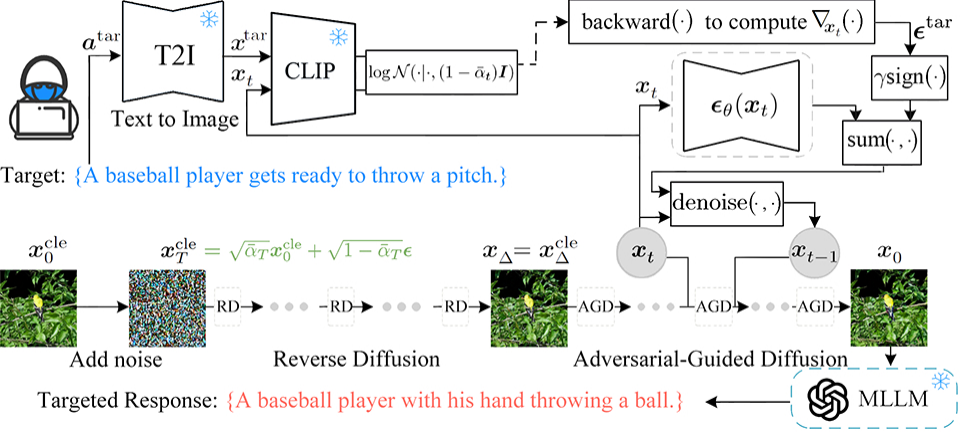}
\caption{Flowchart of the proposed adversarial-guided diffusion process. We perform forward noise addition on the clean image. During the reverse diffusion process, we execute step-by-step denoising from step $T$ to $\Delta$ to reconstruct the original clean image, ensuring the constraint is largely satisfied. In the adversarial-guided diffusion process, we inject target information into noise from $\Delta$ to 1. The resulting image closely approximates the clean image’s structure while injecting the target semantics. }
\label{framework}
\end{figure*}
\section{Related Work}\label{sec:related work}
\textbf{Adversarial attack on MLLMs. }
Adversarial attacks~\cite{goodfellow2014explaining} typically function by introducing imperceptible perturbations into clean images, result in misleading the targeted  responses~\citep{goodfellow2014explaining,kurakin2018adversarial,dong2018boosting}. In the case of MLLMs, robustness of MLLMs is highly dependent on their most vulnerable input visual modality. Attackers can exploit the inherent weaknesses within the model's visual structure to craft adversarial images. Adversarial attacks on MLLMs are generally categorized into two types: black-box~\citep{zhao2023evaluate,dong2023robust,bailey2023image} and white-box attacks~\citep{shayegani2024jailbreak,gao2024boosting,cui2024robustness}. According to the attack objectives, these can further be divided into untargeted~\citep{schlarmann2023adversarial,cui2024robustness} and targeted attacks~\citep{zhao2023evaluate,wang2023instructta,xie2025chain}. 
Compared to the white-box access scenario, the scenario in which adversaries have only black-box access and seek to deceive the model into returning the targeted responses represents the most realistic and high-risk scenario. For the black-box access open-source MLLMs, AttackVLM~\citep{zhao2023evaluate} provides a comprehensive evaluation of the robustness of them to highlights the challenges of targeted adversarial attacks on MLLMs. CoA~\cite{xie2025chain}enhances the generation of adversarial examples by a multi-modal semantic fusion strategy. But this pixel-based attack directly disturbs pixel space easily to be perceived and defended.


\textbf{Diffusion-based unrestricted adversarial attack. }
Due to the $\ell_{p}$-norm distance is inadequate to capture how
human perceive perturbation accurately~\citep{chen2023advdiffuser,shamsabadi2020colorfool,yuan2022natural}, a number of unrestricted attack methods have been proposed to improve pixel-based attack methods. In recent, diffusion models have been introduced into adversarial attack research due to it's~\citep{ho2020denoising,song2021denoising,song2021scorebased} capable of generating natural and diverse outputs. Diffusion-based unrestricted methods such as AdvDiff~\citep{dai2023advdiff} and AdvDiffVLM~\cite{guo2024efficient} incorporate adversarial guidance during the reverse-diffusion process by injecting adversarial gradients, enabling the generation of adversarial images. Similarly, AdvDiffuser~\cite{chen2023advdiffuser} applies Projected Gradient Descent (PGD)~\cite{madry2018towards} within the reverse-diffusion process. However, methods like and AdvDiffuser, which inject adversarial semantics at each step in the reverse-diffusion process, tend to significantly degrade the quality of the generated images. AdvDiffVLM adds adversarial grad directly in the final steps of the denoising process with multiple outer loop for embedding more target semantics. Both diffusion-based attacks use adversarial inpainting to linear interpolation between the image and noise results in poor reconstruction, which coupled with the perturbations strategy, results in suboptimal robustness in quality and attack capability. In addition, ACA~\cite{chen2023contentbased} adversarial perturbations into initial latent image, which leads to substantial deviations in the generated image content due to error accumulation during the reverse-diffusion process.
\section{Preliminaries}
\subsection{Problem definition}
\label{sec3-1}
For an MLLM, a VQA task is defined by
\begin{equation}
\A=f(\x,\q)
\end{equation}
where $\x$ represents an input image, $\q$ is a text question regarding as the content of $\x$, and $\A$ represents both the textual answer and its embedding for notational simplicity, as they are semantically aligned in this context.

In this paper, $\q$ of the MLLM is a placeholder. We focus on targeted adversarial attacks against the MLLM, aiming to generate an adversarial image $\x$ that mislead the MLLM into responding with specific target answer. This objective is defined by
\begin{equation}
\begin{aligned}
\max\,
&\A^\T\A^{\rm tar}\\
{\rm s.t.}\,
&\|\x_0-\x_0^{\rm cle}\|_\infty\leqslant\delta,
\end{aligned}
\label{objective}
\end{equation}
where $\A^{\rm tar}$ is the adversarial target text that the adversary expects the victim models to return, $\x_0$ is the adversarial image, $\x_0^{\rm cle}$ is the clean image, and $\delta$ denotes a threshold constant of controlling adversarial image quality. According to~\citep{goodfellow2014explaining}, adversarial attacks on MLLMs summarized as the optimization of two objectives.

\subsection{Diffusion Model}\label{sec3-2}
As an effective generative model, the diffusion model~\citep{ho2020denoising} has be demonstrated that it generates images of higher quality and diversity than GANs~\citep{dhariwal2021diffusion}. Diffusion models operate by defining a Markov chain and learning a denoising process to sample from a standard normal distribution. This diffusion involves two directions: a forward diffusion and a reverse diffusion.

In forward diffusion, the noisy image $\x_t$ is a combination of the clean image $\x_0$ and a computer-generated noise $\EPS$
\begin{align}
\x_t 
&= \sqrt{\bar{\alpha}_t} \x_0 + (1 - \sqrt{\bar{\alpha}_t}) \EPS, \label{xtx0eps}
\end{align}
where $\bar{\alpha}_t$ is defined as in DDPM~\citep{ho2020denoising}. In reverse diffusion, given $\x_t$, we can approximate $\x_0$ by,
\begin{align}
\x_0=\frac{1}{\sqrt{\bar\alpha_t}}\x_t-\frac{\sqrt{1-\bar\alpha_t}}{\sqrt{\bar\alpha_t}}\EPS_{\theta}(\x_t)\label{x0xteps}
\end{align}
where $\EPS_{\theta}(\cdot)$ denoted the pretrained noise prediction neural network.
Generally, the smaller the value of $t$, the smaller of the approximation error of $\x_0$ by Eq.~\eqref{x0xteps}.

For a well-trained diffusion model, we regard the $\EPS_{\theta}(\x_t)$ follow the normal Gaussian probability distribution,
\begin{align}
\EPS_{\theta}(\x_t)\approx\EPS\sim\mathcal{N}(\bm0,\bm I)\,.
\label{Gauss}
\end{align}





\section{Adversarial-guided diffusion}\label{sec:method}
In this paper, we use Stable Diffusion~\citep{rombach2022high} as our diffusion model.
As shown in Figure~\ref{framework}, we propose an iterative adversarial-guided diffusion (AGD) for generating adversarial images by injecting targeted information. 

\subsection{Reverse diffusion for the constraint in Eq.~\eqref{objective}.} The constraint in Eq.~\eqref{objective} is that the image must visually resemble the original clean image. To enforce this constraint, we add noise directly to the clean image during the forward diffusion process. 
In the reverse denoising diffusion, adversarial image generation begins with the noised version of the clean image at time step $T$, such that $\x_T^{\rm cle}=\sqrt{\bar{\alpha}_T}\x^{\rm cle}_0+\sqrt{1-\bar{\alpha}_T}\EPS\,$. 
Starting from $\x_T^{\rm cle}$, standard reverse sampling is applied to obtain $\x_\Delta^{\rm cle}$. We set $\Delta$ to a very small value, ensuring that, as shown in Figure~\ref{framework}, the constraint in Eq.~\eqref{objective} is essentially satisfied.
Each reverse-diffusion step is given by
\begin{align}
\x_{t-1}
&=\bm\mu_t+\sqrt{\beta_t}\bm e_t\label{ddim1}\\
\bm\mu_t
&=\frac1{\sqrt{1-\beta_t}}\left\{\x_t-\frac{\beta_t}{\sqrt{1-\bar{\alpha}_t}}
\EPS_{\theta}(\x_t)
\right\}
\label{ddim2}
\end{align}
where $\theta$ represents the pretrained parameters of the diffusion model, $\beta_t$ is defined as in DDPM~\citep{ho2020denoising} and $\bm e_t$ is a predefined noise image by its clean image $\x_0^{\rm cle}$~\citep{huberman2024edit}. In practice, we input $\x_T^{\rm cle}$ and a prompt into SD, where the prompt is generated by BLIP-2~\cite{li2023blip} based on the corresponding clean image. For simplicity, we omitted the prompt in $\EPS_{\theta}(\x_t)$.

Using this specific $\bm e_t$ instead of standard Gaussian makes $\x_{t-1}$ more deterministic, reducing randomness, and ensures that ${\x}^{\rm cle}_\Delta$ after the $\Delta$ step is more similar to ${\x}_0^{\rm cle}$. The details of how to obtain $\bm e_t$ is provided in~~\citep{huberman2024edit}.

The reverse diffusion uses $\x_t$ and the pretrained $\theta$ to infer the noise $\EPS_{\theta}(\x_t)$, which is subtracted from $\x_t$ to obtain $\x_{t-1}$. This step constitutes the denoising process. By combining Eqs~\eqref{ddim1} and \eqref{ddim2}, we define a denoising function:
\begin{align}
\x_{t-1}= {\rm denoise}\bigl(\x_t,\EPS_{\theta}(\x_t)\bigr)\,.
\label{denoise}
\end{align}

As shown in Figure~\ref{framework}, during the reverse diffusion from $T$ down to $\Delta$, $\x_{t-1}$ is step-by-step derived from $\x_t$ using Eq.~\eqref{denoise}. This ensures that the constraint of Eq.~\eqref{objective} is largely satisfied, meaning that the image, ${\x}^{\rm cle}_\Delta$, is essentially similar to the original clean image, i.e., ${\x}^{\rm cle}_\Delta \approx {\x}_0^{\rm cle}$.

\subsection{AGD for the objective of Eq.~\eqref{objective}.}
For injecting the target information, we assign $\x_\Delta^{\rm cle}$ to $\x_\Delta$. From $\Delta$ to $1$, we use AGD steps. First, a target image $\x_0^{\rm tar}$ is generated by $\x_0^{\rm{tar}}={\rm T2I}(\A^{\rm{tar}})$ from the target text $\A^{\rm{tar}}$ by a text-to-image model~\cite{rombach2022high}.

 In general, given ${\x}_t$ and its corresponding $\EPS_{\theta}(\x_t)$, we can iteratively inject the target information into the starting image using the denoising process by Eq.~\eqref{denoise}. We apply the effect shown in Eqs.~\eqref{denoise} and \eqref{x0xteps}. Given $\epsilon^{\rm tar}$, we can step-by-step inject the target information from the injection starting step $\x_\Delta$. The injection noise is denoted by $\EPS^{\rm tar}$, it is integrated into the existing noise term by
$\tilde{\EPS}=\EPS_{\theta}(\x_t)+\EPS^{\rm tar}$\,.

From Eq.~\eqref{xtx0eps}, we have 
\begin{align}
\x_t\sim p(\x_t|\x_0^{\rm tar})=\mathcal N(\x_t|\sqrt{\bar{\alpha}_t}\x_0^{\rm tar},(1-\bar{\alpha}_t)\bm I)
\label{xtdist}
\end{align}

With the Tweedie function~\citep{song2021scorebased}, $\EPS^{\rm tar}$ can be calculated from the Gaussian distribution,
\begin{align}
\EPS^{\rm tar}
&=-\sqrt{1-\bar{\alpha}_t}
\nabla_{\x_t}\log p(\x_t|\x_0^{\rm tar})\,.
\label{advnoise}
\end{align}
When $t$ is small, causing $\bar\alpha_t$ to be close to 1, taking the logarithm of the Gaussian function in Eq.~\eqref{xtdist} results in a similarity measure between $\x_t$ and $\x_0^{\rm tar}$. To better inject target semantic information, we compute this similarity using CLIP~\citep{radford2021learning} features. Specifically, we replace $\x_t$ and $\x_0^{\rm tar}$ in Eq.~\eqref{xtdist} with their normalized CLIP features.

To enhance the adversarial attack effect, we apply a 
${\rm sign}$ function to injection noise and tune the contributions of the two terms. The adversarial-guided noise is defined by
\begin{align}
\tilde{\EPS}=\EPS_\theta(\x_t) 
+\gamma{\rm sign} (\EPS^{\rm tar})\,,
\label{finalnoise}
\end{align}
where $\gamma$ is scale factor to control target semantic, $\rm sign(\cdot)$ outputs positive or negative one, based on the input's sign.

In a well-trained diffusion model, the noise term follows a Gaussian distribution as shown in Eq.~\eqref{Gauss}, and the noise term generated from this distribution has equal power across all frequencies, effectively covering the entire frequency domain. This method of injecting target information using Gaussian noise is generally difficult to defend against, particularly when using low-pass filtering (see experimental evidence in \S\ref{defense}). As shown in Eq.~\eqref{xtx0eps}, during the diffusion steps from $\Delta$ to 1, $\x_t$ consists of two components. When applying low-pass filtering to $\x_t$, the second component spans the entire frequency domain, making it difficult to filter out. This is exactly where we inject the target information in the form of Gaussian noise.

\begin{algorithm}[!t]
\caption{Adversarial image generation algorithm.}
\begin{algorithmic}[1]
\REQUIRE $\A^{\rm tar}, \EPS_{\theta}(\x)$, $\Delta$, and $\x_0^{\rm cle}$;
\STATE $\x_0^{\rm tar}={\rm T2I}(\A^{\rm tar})$\,;
\STATE $\z_0^{\rm tar}={\rm CLIP}(\x_0^{\rm tar})$\,;
\STATE Obtain $\x_T^{\rm cle}=\sqrt{\bar{\alpha}_T}\x^{\rm cle}_0+\sqrt{1-\bar{\alpha}_T}\EPS\,,$ $\EPS\sim\mathcal N(\bm0,\bm I)$\,;
\FOR{$t\in\{T,\dots,2,1\}$}
\IF{$t>\Delta$}
\STATE $\x^{\rm cle}_{t-1}
= {\rm denoise}(\x^{\rm cle}_t,\EPS_{\theta}(\x_t^{\rm cle}))$\,;
\ELSE 
\IF{$t==\Delta$}
\STATE $\x_t=\x^{\rm cle}_t$
\ENDIF
\STATE $\tilde{\x}=\x_t; \EPS=\EPS_{\theta}(\x_t); \overline{\EPS}=0$\,;
\FOR{$\tau\in\{N,\dots,2,1\}$}
\STATE  $\EPS
=\EPS
+\gamma{\rm sign}( \overline{\EPS})$\,;
\STATE $\overline{\EPS}=\lambda\overline{\EPS}+(1-\lambda)\EPS^{\rm tar}$\,;
\STATE $\tilde{\x}= {\rm denoise}(\tilde{\x},\EPS)$\,;
\STATE $\z={\rm CLIP}(\tilde{\x})$\,;
\STATE $\EPS^{\rm tar}=-\sqrt{1-\bar\alpha_{\tau}}\nabla\mathcal N(\z|\sqrt{\bar{\alpha}_{\tau}}\z_0^{\rm tar},(1-\bar{\alpha}_{\tau})\bm I)$
\ENDFOR
\STATE $\tilde{\EPS}
=\EPS_{\theta}(\x_t) 
+\gamma
{\rm sign}(
\overline{\EPS})$\,;
\STATE $\x_{t-1}= {\rm denoise}(\x_t,\tilde{\EPS})$\,;
\ENDIF
\ENDFOR
\RETURN An adversarial image: $\x_0$\,.
\end{algorithmic}
\label{algshort}
\end{algorithm}

\begin{figure}[!t]
\centering
\includegraphics[width=0.4\textwidth]{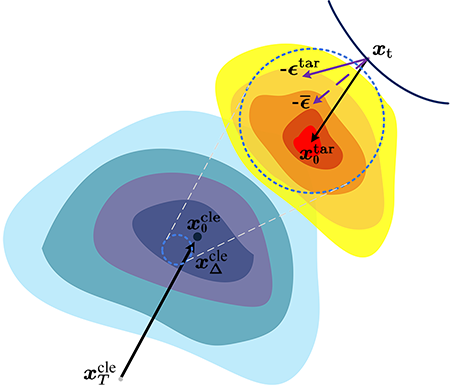}
\caption{
At each time step $t$ of AGD, we use the effect shown in Eq.~\eqref{x0xteps}: $\x_0=\frac{1}{\sqrt{\bar\alpha_t}}\x_t-\frac{\sqrt{1-\bar\alpha_t}}{\sqrt{\bar\alpha_t}}\bar{\EPS}$, to inject the target information to current $\x_t$.
}
\label{geometric}
\end{figure}
\begin{table*}[!t]
\centering
    \resizebox{0.8\linewidth}{!}{
\begin{tabular}{l|l||cccccc|c}
            \hline
            \rowcolor{mygray}&
&\multicolumn{6}{c}{Text encoder (pretrained) for evaluation}\vline
&\\
\cline{3-8}
\rowcolor{mygray}\multirow{-2}{*}{\textbf{MLLM}}  & \multirow{-2}{*}{\textbf{Method}}  &\textbf{RN50} & \textbf{RN101} & \textbf{ViT-B/16} & \textbf{ViT-B/32} & \textbf{ViT-L/14} & \textbf{Ensemble} &\multirow{-2}{*}{\textbf{ASR }}\\ 
\hline\hline
\multirow{6}{*}{UniDiffuser}  
& MF-it& 0.655& 0.639 & 0.670 & 0.698 & 0.611    & 0.656  & $80.9\%$    \\                         
& MF-ii& 0.709& 0.695 & 0.722 & 0.733 & 0.637    & 0.700 &$90.7\%$    \\                    
& AdvDiffuser &0.499 & 0.486& 0.453& 0.472& 0.338& 0.424&$37.2\%$\\
& ACA&0.448&0.439&0.456&0.466&0.322&0.426&$35.7\%$ \\
& AdvDiffVLM&0.670 & 0.656&0.685 &0.702 & 0.597&0.662 &84.4\%\\
& CoA&0.698 &0.682 & 0.712 & 0.729 & 0.619 & 0.688 &  88.2\%\\
& AGD (ours) &   \textbf{0.731}     &  \textbf{0.718} &    \textbf{0.742}   &      \textbf{0.755}       &     \textbf{0.663}     &\textbf{0.721} & $\textbf{95.4\%}$\\
\hline
\multirow{6}{*}{BLIP-2}  
& MF-it& 0.492& 0.474 & 0.520& 0.546& 0.384& 0.483&$84.2\%$ \\
& MF-ii& 0.562& 0.541 & 0.573& 0.592& 0.449& 0.543&$87.9\%$  \\          
& AdvDiffuser&0.493&0.471&0.505&0.529&0.384&0.476&$29.5\%$\\
& ACA &0.472&0.458&0.478&0.458&0.349&0.450&$25.9\%$\\
& AdvDiffVLM& 0.551&0.529 & 0.572& 0.596&0.459 &0.541 &66.2\%\\
& CoA& 0.428& 0.421 & 0.437 & 0.456 & 0.314&0.411 &  22.6\%\\
& AGD (ours) &\textbf{0.725} &\textbf{0.710} &\textbf{0.736} &\textbf{0.747} &\textbf{0.652}&\textbf{0.714}& $\textbf{89.4\%}$\\
\hline
\multirow{6}{*}{LLaVA-1.5} 
& MF-it& 0.389& 0.441 &0.417& 0.452& 0.288& 0.397&$32.7\%$ \\
& MF-ii& 0.396& 0.440 &0.421&0.450& 0.292& 0.400&$34.1\%$ \\   
& AdvDiffuser&0.501&0.484&0.505&0.513&0.364&0.473&$17.9\%$\\
& ACA &0.496&0.482&0.500&0.509&0.358&0.469&$17.1\%$\\
& AdvDiffVLM&0.528 &0.512 &0.546 &0.562 &0.420 &0.513 &36.4\%\\
& CoA& 0.481&0.480 & 0.494 &  0.490&0.346  & 0.458 &  16.2\%\\
& AGD (ours) &\textbf{0.554}&\textbf{0.543}&\textbf{0.558}&\textbf{0.565}&\textbf{0.428}&\textbf{0.529}&$\textbf{41.8\%}$\\
\hline
\multirow{6}{*}{MiniGPT-4 }  
& MF-it& 0.472& 0.450& 0.461& 0.484& 0.349& 0.443&$74.8\%$\\  
& MF-ii& 0.525&0.541& 0.542& 0.572& 0.430& 0.522&$76.1\%$\\                           
& AdvDiffuser&0.378&0.365&0.417&0.434&0.288&0.376&$28.8\%$\\
& ACA&0.428&0.380&0.457&0.463&0.306&0.407&$31.1\%$\\
& AdvDiffVLM& 0.500&0.414 & 0.531& 0.559&0.403 &0.482 &57.4\%\\
& CoA&0.442 &0.354 &0.463  &  0.471&0.323  &0.410  &  35.2\%\\
& AGD (ours)&\textbf{0.674}&\textbf{0.594}&\textbf{0.688}&\textbf{0.704}&\textbf{0.579}&\textbf{0.648}& $\textbf{88.0\%}$\\
\hline
\multirow{6}{*}{Qwen2-VL} 
& MF-it  &0.286&0.398&0.324&0.315&0.214&0.307&$23.8\%$  \\    
& MF-ii &0.281&0.399&0.327&0.316&0.217&0.309&$24.2\%$\\ 
& AdvDiffuser&0.267&0.395&0.316&0.302&0.198&0.196&7.6\%\\
& ACA &0.257&0.400&0.299&0.292&0.187&0.288&4.7\%\\
& AdvDiffVLM&0.345 &0.424 &0.383 &0.393 & 0.256&0.360 &16.4\%\\
& CoA& 0.269& 0.391& 0.307 & 0.297 & 0.196 & 0.292 &  12.4\%\\
&AGD (ours)  & \textbf{0.372}& \textbf{0.467} & \textbf{0.414} &\textbf{0.409} & \textbf{ 0.276}&\textbf{0.387}&$\textbf{30.5\%}$\\
\hline
\end{tabular}
}
\caption{Comparing AGD with SOTA adversarial attack methods for targeted attacks against victim MLLMs. We report the ASR $\uparrow$ and CLIP score $\uparrow$ computed by different CLIP text encoders and their ensemble/average results. The best result is bolded.}
\label{attackresult}
\end{table*}

\textbf{Momentum-based injection.} Now, focusing on the yellow contours in Figure~\ref{geometric}, we aim to inject the target information to the adversarial image $\x_{t}$. The details of the momentum-based injection algorithm is shown in Algorithm~\ref{algshort}. In Algorithm~\ref{algshort}, the AGD ensures that the diffusion direction gradually approaches the target denoising direction. Similar to the gradient descent, we introduce momentum $\overline{\EPS}$ to accelerate the guidance towards the target direction by using an Exponential Moving Average (EMA). The momentum-based noise also aligns with the smaller time step, and an additional advantage is that momentum gradients easily converge to  $\x^{\rm tar}$. As momentum $\overline{\EPS}$ moves in a more optimal fit direction, shown in Figure~\ref{geometric}, the $\EPS^{\rm tar}$ approaches the optimal gradient direction. EMA is given by
\begin{equation}
\begin{aligned}
&\overline{\EPS}=\lambda \overline{\EPS}+(1-\lambda)\EPS^{\rm tar}
\end{aligned}
\end{equation}
where $\overline{\EPS}$ is the accumulated ${\EPS}^{tar}$, with an initial value of 0, $\lambda$ denotes momentum factor and $\lambda\in(0,1)$, with larger $\lambda$ resulting in less volatile changes of the momentum.

Two inequalities from DDPM~\cite{ning2024elucidating,li2024alleviating} can be used to guide the addition of target information to the $\x_{t}$. The two inequalities~\cite{ning2024elucidating,li2024alleviating} about momentum iteration step $\tau$  are given by, 
\begin{align}
\|\EPS_{\theta}(\x_{\tau}^{\rm tar})\|
&<
\|\EPS_{\theta}(\x_{\tau})\|\label{xtar}\\
\|\EPS_{\theta}(\x_{\tau-1})\|
&<
\|\EPS_{\theta}(\x_{\tau})\|\,.
\label{correct}
\end{align}
In the reverse-diffusion, if we use $\x_{\tau}^{\rm tar}$ to simulate the denoising process~\cite{ning2024elucidating,li2024alleviating}, we have the inequality \eqref{xtar}. 
It is preferable to fit $\EPS_{\theta}(\x_{\tau}^{\rm tar})$, so we achieve this by using $\EPS_{\theta}(\x_{\tau-1})$ to instead of $\EPS_{\theta}(\x_{\tau})$. By using \eqref{xtar} and \eqref{correct}, this approach allows us to use the noise from a smaller time step to replace the noise at the current time step~\cite{ning2024elucidating,li2024alleviating}. To implement this, we calculate an EMA of the noise from several smaller time step, substituting noise at the current step.

\section{Experiment}\label{sec:experiment}
\begin{table*}[h]
  \centering
    \resizebox{\linewidth}{!}{
		\setlength\tabcolsep{5pt}
		\renewcommand\arraystretch{1}
        \begin{tabular}{l||ccccccccc}
            \hline
            \rowcolor{mygray}
&\multicolumn{3}{c}{UniDiffuser}  &\multicolumn{3}{c}{BLIP-2}&\multicolumn{3}{c}{LLaVA-1.5}\\
            \cline{2-10}
            \rowcolor{mygray} \multirow{-2}{*}{\textbf{Method}}& \textbf{SSIM $\uparrow$} & \textbf{LPIPS $\downarrow$}  & $\rm{\mathbf{PSNR}}_{\times 10^{2}}$  $\uparrow$&\textbf{SSIM $\uparrow$} & \textbf{LPIPS $\downarrow$}  & $\rm{\mathbf{PSNR}}_{\times 10^{2}}$  $\uparrow$&\textbf{SSIM $\uparrow$} & \textbf{LPIPS $\downarrow$}  & $\rm{\mathbf{PSNR}}_{\times 10^{2}}$ $\uparrow$ \\
            \hline \hline
MF-it  & 0.6675  & 0.3502  &  0.2580 &0.6841 &  0.3341 &     0.2607   & 0.6813 &0.3349 &  0.2605\\
    MF-ii  & 0.6705  &  0.3435 & 0.2583  & 0.6834 &   0.3352&   0.2605    & 0.6817& 0.3338& 0.2605  \\
AdvDiffuser   & 0.3541  &  0.5081 &  0.1329 &0.3494& 0.5228&  0.1333   & 0.4489&  0.5080 & 0.1327  \\
  ACA  & 0.5872 & 0.4558  & 0.1821 & 0.5946  &  0.4461 &  0.1835 &0.6052& 0.4367&  0.1858     \\
    AdvDiffVLM  & 0.6062 & 0.3717  & 0.2134 & 0.6062 & 0.3717  & 0.2134 &0.6062 & 0.3717  & 0.2134    \\
    CoA  & 0.6193 & 0.2176 & 0.2512 &  0.6193& 0.2176 & 0.2512 &  0.6193 & 0.2512  &   0.2176  \\
   AGD (ours) & \textbf{0.8351}  &  \textbf{0.1097}  & \textbf{0.2696} & \textbf{0.8494}  &  \textbf{0.0935}  &   \textbf{0.2766}  &\textbf{0.8352} & \textbf{0.1094}& \textbf{0.2696}\\
            \hline
        \end{tabular}}  
    \caption{Performance comparison of different adversarial attack methods evaluated using SSIM, LPIPS, and PSNR metrics. Higher SSIM and PSNR values indicate better structural and perceptual similarity to the original image, while lower LPIPS values indicate better visual similarity. This comparison highlights each method’s difference in image visual quality.}
\label{imagequality}
\end{table*}

 \begin{table}[!t]
\centering
\resizebox{0.99\linewidth}{!}{
\setlength\tabcolsep{4pt}
\renewcommand\arraystretch{1}
\begin{tabular}{l|l||cccc}
\hline
\rowcolor{mygray}Defenses&MLLM  &Unidiffuser&LLaVA-1.5&MiniGPT-4&BLIP-2\\

\hline \hline
\multirow{4}{*}{JEPG}
&MF-it & $30.7\%$ & $19.7\%$ & $19.0\%$ & $25.5\%$ \\
&MF-ii & $38.1\%$ & $21.1\%$ & $24.9\%$ & $25.6\%$ \\
&AdvDiffVLM&66.2\% & 29.4\% & 40.6\% & 45.8\%\\
&AGD & $\textbf{92.1\%}$ & $\textbf{36.1\%}$ & $\textbf{82.4\%}$ & $\textbf{87.6\%}$\\

\cline{1-6}
\multirow{4}{*}{RP} 
&MF-it& $26.2\%$& $17.0\%$ & $26.0\%$ & $23.4\%$\\
&MF-ii& $27.8\%$& $16.3\%$ &  $26.3\%$ &$26.5\%$\\
&AdvDiffVLM&50.4\% &   20.5\%& 42.8\% & 44.0\%\\
&AGD& $\textbf{57.5\%}$& $\textbf{21.6\%}$ & $\textbf{57.2\%}$ &$\textbf{60.8\%}$\\
\cline{1-6}
\multirow{4}{*}{SOAP} 
&MF-it&$62.5\%$&$22.0\%$  &  $63.6\%$& $71.0\%$\\
&MF-ii&$85.2\%$ & $23.5\%$ & $68.3\%$ &$68.0\%$\\
&AdvDiffVLM& 81.0\%& 33.4\% & 48.6\% & 53.1\%\\
&AGD&$\textbf{91.3\%}$ & $\textbf{36.3\%}$ & $\textbf{76.8\%}$ &$\textbf{88.5\%}$\\
\cline{1-6}
\multirow{4}{*}{DiffPure} 
&MF-it& $27.1\%$ & $17.3\%$ & $23.1\%$ & $25.8\%$ \\
&MF-ii & $29.7\%$ & $16.6\%$ & $25.6\%$ & $26.1\%$ \\
&AdvDiffVLM& 70.4\%& 30.4\% &49.2\%  &45.8\% \\
&AGD & $\textbf{89.4\%}$ & $\textbf{31.4\%}$ & $\textbf{80.8\%}$ & $\textbf{82.3\%}$\\
\cline{1-6}
\multirow{4}{*}{MimicDiffusion} 
&MF-it&$25.8\%$ & $15.9\%$ & $26.5\%$ & $24.4\%$\\
&MF-ii&$28.3\%$ & $16.2\%$ & $28.2\%$ &$24.8\%$\\
&AdvDiffVLM&46.4\% & $\textbf{21.4\%}$ &  43.2\%& 39.4\%\\
&AGD&$\textbf{72.6\%}$ & $21.0\%$ &$\textbf{50.8\%}$ & $\textbf{51.2\%}$ \\
\hline
\end{tabular}}
\caption{Attack performance against adversarial defenses. We report ASR results on five adversarial defense methods compared with baselines.}
\label{defense_asr}
\end{table} 

\begin{table}[!t]
\centering
\setlength\tabcolsep{5pt}
\renewcommand\arraystretch{1}
\resizebox{0.46\textwidth}{!}{\begin{tabular}{c||cc|cc|cc}
\hline
 \rowcolor{mygray}  &  \multicolumn{2}{c|}{UniDiffuser}& \multicolumn{2}{c|}{BLIP-2} &\multicolumn{2}{c}{MiniGPT-4} \\
 \cline{2-7}
\rowcolor{mygray}\multirow{-2}{*}{${\bm N}$}&\textbf{ASR} $\uparrow$&\textbf{LPIPS $\downarrow$}&\textbf{ASR}$\uparrow$ & \textbf{LPIPS $\downarrow$}&\textbf{ASR}$\uparrow$&\textbf{LPIPS $\downarrow$}  \\ 
 \hline
 \hline
1&67.9\%&0.145&38.9\%&0.142&39.1\%&0.142    \\               
10&$83.6\%$&0.127&46.7\%&0.120&42.6\%&0.120   \\               
30&88.6\%&0.115&67.2\%&0.104&$64.3\%$&0.104\\
50&$\textbf{95.4\%}$&\textbf{0.109}&$\textbf{89.4\%}$&\textbf{0.093}&$\textbf{88.0\%}$&\textbf{0.093}\\
\hline
\end{tabular}}
\caption{Ablation study of iterative denoising steps in AGD. We report ensemble CLIP score and LPIPS , $N=1$ indicates without momentum-based injection.}
\label{ablation1}
\end{table}
 
\textbf{Datasets.} 
The dataset consists of both images and prompts. Following~\citep{zhao2023evaluate}, We use validation set of ImageNet-1K~\cite{deng2009imagenet} as clean images, and we randomly select 1000 text
descriptions from from MS-COCO captions~\citep{lin2014microsoft} as our target texts. 

\noindent\textbf{Victim MLLMs.}
To evaluate the performance of our AGD on the MLLMs attack, We conducted targeted adversarial attack experiments on several advanced open-source MLLMs, including UniDiffuser~\citep{bao2023one}, which employs a diffusion-based framework to jointly model the distribution of image-text pairs.
BLIP-2~\citep{li2023blip}, integrates a querying transformer and a large language model to boost image-grounded text generation. Furthermore, MiniGPT-4~\citep{zhu2023minigpt} and LLaVA-1.5~\citep{liu2024visual} have scaled up the capabilities of large language models, utilizing Vicuna-13B~\citep{chiang2023vicuna} as language model. Qwen2-VL~\cite{wang2024qwen2} enhances multimodal processing with dynamic-resolution image processing and unified modeling.

\noindent\textbf{Baselines.}
To evaluate the performance of our method, we will compare it with existing attack methods in SOTA multimodal large models with gray-box setting, including MF-it~\citep{zhao2023evaluate}, MF-ii~\citep{zhao2023evaluate}, and CoA~\cite{xie2025chain}, and SOTA adversarial attack method based on diffusion model AdvDiffuser~\citep{chen2023advdiffuser}, ACA~\citep{chen2023contentbased}, and AdvDiffVLM~\cite{guo2024efficient}.

\noindent\textbf{Evaluation metrics.}
Following~\citep{zhao2023evaluate}, we adopt CLIP score~\citep{hessel2021clipscore}, which compute the text similarity between response text generated by the victim models and target texts using different CLIP text encoder. Moreover, we also adopt attack success rate (ASR) to evaluate attack performance. To assess the quality of adversarial images, we employ three evaluation metrics: SSIM~\citep{wang2004image}, LPIPS~\citep{zhang2018unreasonable}, and PSNR~\citep{hore2010image}.

\noindent\textbf{Experimental Details.}
We use clean images to generate adversarial images with fixed resolution 512. We set scale parameter $\gamma=6$, the number of iteration $N=50$, momentum factor $\lambda=0.9$, and $\Delta=5$. In addition, we use Stable Diffusion~\citep{rombach2022high} (the number of forward diffusion steps $T = 100$) to generate adversarial images.
\subsection{Targeted attack results on MLLMs}
As shown in Tabel~\ref{attackresult}, we evaluate effectiveness of our method on different victim MLLMs. Compared with recent targeted adversarial attack methods: MF-ii, MF-it and CoA, diffusion-based method AdvDiffuser, ACA, and AdvDiffVLM, experiment results demonstrate that our method consistently outperforms baselines in terms of CLIP score and ASR. Specifically, our method exhibit significant improvements of targeted attack such as UniDiffuser and BLIP-2. Indicating the effectiveness of our methods targeted attack against victim MLLMs. For MiniGPT-4, the generated target images sometimes contain extraneous content, which can lead to a decrease in the similarity between MLLM's response and the target text.
\subsection{Visualization}
\begin{figure*}[!t]
\centering
\includegraphics[width=0.95\textwidth]{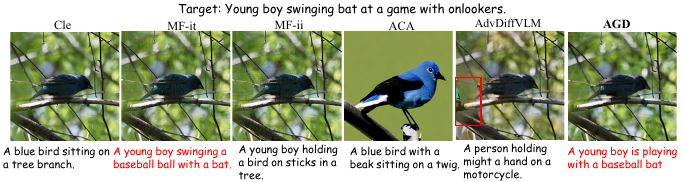}
\caption{\textbf{Comparison of targeted adversarial attacks results on UniDiffuser.}  Adversarial target text appears above each image, and captioning results from clean or adversarial images are shown below. }
    \label{visual}
\end{figure*}
 \begin{figure}[!t]
    \centering
\subfloat[Scale $\gamma$]
{
\includegraphics[width=0.200\textwidth]{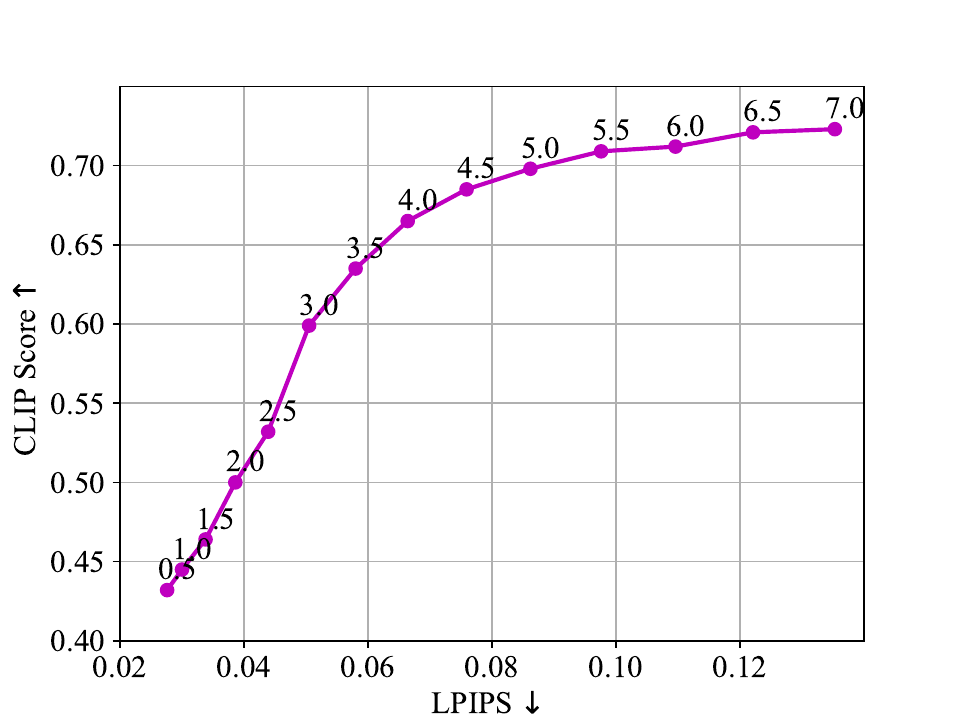}
\label{scale}
}
\subfloat[Hyperparameter $\Delta$]
{
\includegraphics[width=0.200\textwidth]{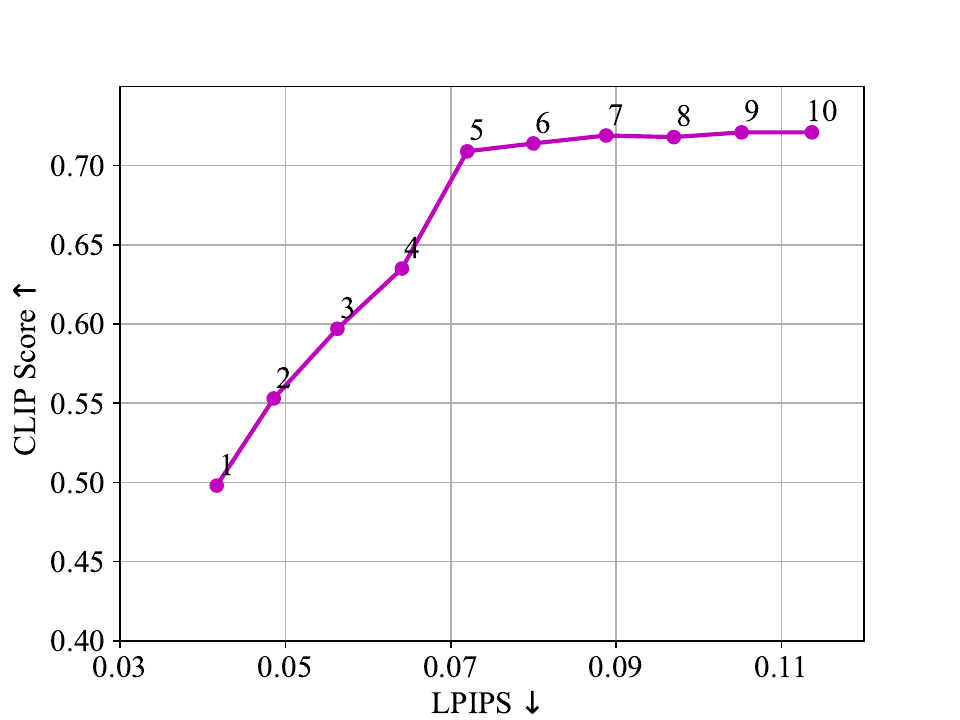}
\label{sample_step}
}
\quad
\subfloat[Inner iterations $N$]
{
\includegraphics[width=0.200\textwidth]{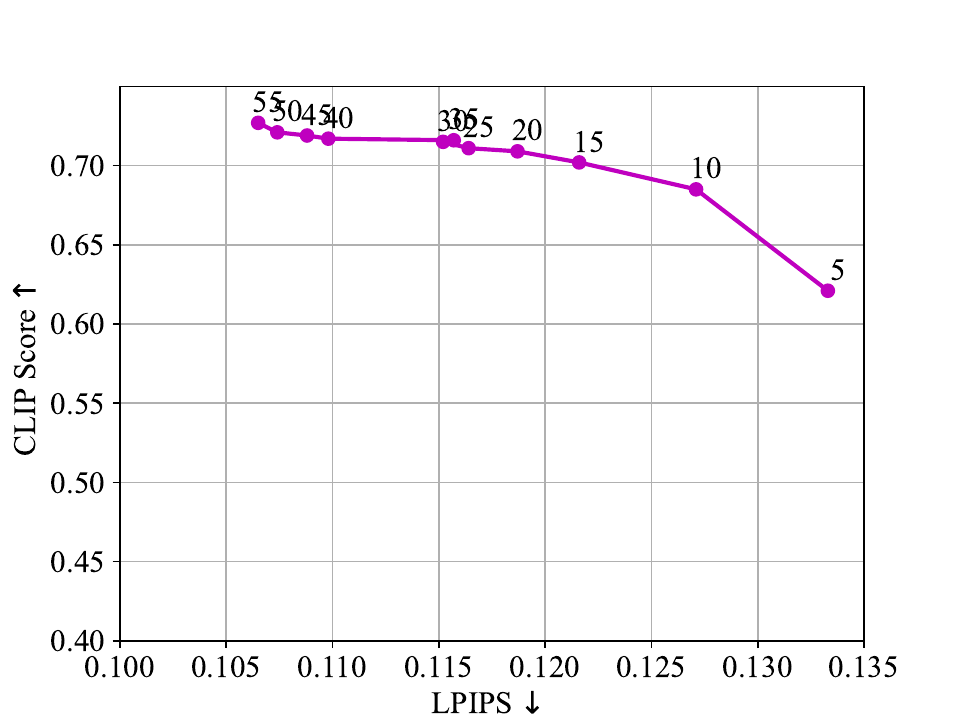}
\label{step}
}
\subfloat[Momentum factor $\lambda$]
{
\includegraphics[width=0.200\textwidth]{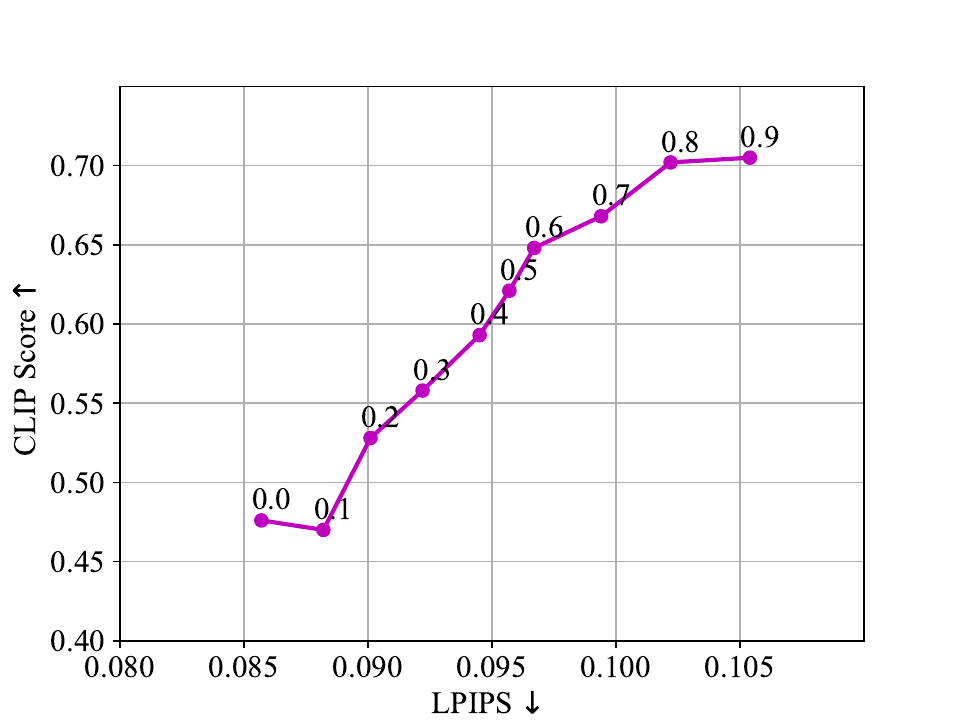}
\label{momentumfactor}
}

\label{ablation}
\caption{Ablation study on the impact of hyperparameters in UniDiffuser.}
\end{figure}
\textbf{Quantitative comparison.}
To evaluate the image quality of adversarial images generated by our method, we quantitatively assess the image quality using image quality
evaluation metrics such as SSIM, LPIPS, and PNSR for the adversarial images in Tabel~\ref{attackresult}. As illustrated in Table~\ref{imagequality}, compared to baselines, the adversarial images generated by our method exhibit higher image quality, especially on LPIPS. This is attributable to the fact that we apply adversarial noise guidance while intelligently choosing the steps for adversarial semantics injection during the reverse-diffusion process, based on the concept of truncated diffusion.




\noindent\textbf{Qualitative comparison.}
We visualize adversarial images generated by our method and other bashlines. As shown in Figure~\ref{visual}, compared to the adversarial images generated by baselines, our method noticeably preserves the structure and natural appearance of the clean images. In contrast, MF-it, MF-ii and CoA directly introduce adversarial perturbations in terms of $\ell_p$-norm limitation to the clean images. ACA significantly changes the image structure by introducing adversarial perturbations to latent during the early stages of the reverse process. Furthermore, AdvDiffVLM introduce extra information of target into reconstruction iamge (red box in the figure). Moreover, we present the responses from MLLMs when input adversarial images are generated by different methods, demonstrating that our method successfully misleads the MLLM's response. 


\subsection{Against adversarial defense models}\label{defense}
To evaluate effectivness of our methods aginst adversarial defense methods, we conduct conduct adversarial defense experiments on various defenses such as JEPG~\cite{guo2018countering}, R$\&$P~\cite{xie2018mitigating}, SOAP~\cite{shi2021online}, DiffPure~\cite{nie2022DiffPure}, and MimicDiffusion~\cite{Song_2024_CVPR}, then we evaluate attack performance as our setting. In Table~\ref{defense_asr}, our method outperforms baselines on adversarial defense methods, demonstrating our method's superiority against adversarial defense. This is because AGD embeds target across the all frequency bands due to our design, unlike conventional methods that focus on high-frequency components. They are more vulnerable to defense that smooths out adversarial perturbations.

\subsection{Ablation study}




\textbf{The impacts of steps in momentum-based injection.} To validate our momentum-based injection strategy, we conduct an ablation study of iterations $N$. As shown in Table~\ref{ablation1}, $N=1$ denotes no momentum-based injection. We observe that adversarial-guided diffusion with momentum-based injection significantly improves the attack performance against MLLMs, indicating its essential role for attack performance and also contributes to boosting imperceptibility.


\noindent\textbf{The impacts of hyperparameters.}
We first explore the impact of hyperparameter adversarial scale $\gamma$, inner iterations $N$, and momentum factor $\lambda$ on the UniDiffuser. We explore impact of $\gamma$  in a range of [0.5, 7.0] with 0.5 intervals, other hyperparameters are the same as the above targeted attack experiments. As shown in Figure~\ref{scale}, the results show that increasing $\gamma$ enhances attack performance but diminishes the visual quality of adversarial images, our choice of hyperparameter considers attack performance and image quality trade-off. Similarly, We conduct experiments with $N$ varies in a range of $[0, 55]$ with 5 intervals and $\lambda$ in a range of $[0, 0.9]$ with 0.1 intervals. From the results in Figure~\ref{step}, we find that larger values for $N$ result in a greater CLIP score, but it does not degrade the quality of adversarial images. This is because, in the inner loop, AGD search optimal adversarial direction, ensuring attack efficacy while minimizing perturbations. From the Figure~\ref{momentumfactor}, bigger momentum factor $\lambda$ results in greater attack performance due to it average a better adversarial guidance. We also explore the effects of the sampling strategy, as shown in Figure~\ref{sample_step}. The results demonstrate that the attack performance improves as $\Delta$ increases. This is attributed to the more adversarial guidance during the reverse-diffusion process, we choose  
$\Delta=5$ to achieve visual imperceptibility and attack performance trade-off.
\section{Conclusion}
In this paper, we introduce AGD, a novel diffusion-based framework for targeted adversarial attacks on MLLMs. During the reverse-diffusion process, we avoid modifying the original clean image's reconstruction to meet attack constraints and ensure visual imperceptibility. In the final steps, we inject target information into full-spectrum noise by using a momentum-based adversarial-guided diffusion process. This guides the adversarial image toward the target, which helps AGD maintain excellent adversarial effectiveness and performance against defenses specifically designed to remove high-frequency perturbations.

{
\small
\bibliographystyle{ieeenat_fullname}
\bibliography{iccv2025}
}
\end{document}